\documentclass{article}

\PassOptionsToPackage{numbers, compress}{natbib}
\usepackage[preprint]{neurips_2023}

\usepackage[utf8]{inputenc} 
\usepackage[T1]{fontenc}    
\usepackage{hyperref}       
\usepackage{url}            
\usepackage{booktabs}       
\usepackage{amsfonts}       
\usepackage{nicefrac}       
\usepackage{microtype}      
\usepackage{xcolor}         
\usepackage{amsmath}
\usepackage{amssymb}
\usepackage{mathtools}
\usepackage{amsthm}
\usepackage{multirow}
\usepackage{wrapfig}
\usepackage[capitalize,noabbrev]{cleveref}
\usepackage{graphicx}
\usepackage{subfigure} 
\usepackage{float}
\usepackage{algorithm} 
\usepackage{algorithmic}
\usepackage{textcomp} 
\usepackage{listings} 
\usepackage{enumitem}
\usepackage{utfsym}
\usepackage{times}
\usepackage{latexsym}
\usepackage{fancyvrb}
\usepackage{CJKutf8}

\newcommand{\commentout}[1]{}

\usepackage{tcolorbox}
\tcbuselibrary{listings}
\definecolor{lightgray}{rgb}{0.95,0.95,0.95}

\newtcblisting{cardlisting}[2][]{colback=lightgray!100!white,
    colframe=gray!80!black,
    arc=4pt,
    boxrule=0.8pt,
    left=4pt,
    right=4pt,
    top=4pt,
    bottom=4pt,
    fonttitle=\bfseries,
    title={#2},
    #1,
    listing only}

\title{WEST: LLM based Speech Toolkit for Speech Understanding, Generation, and Interaction}

\author{
\normalfont{
    Binbin Zhang\textsuperscript{4,5}, Chengdong Liang\textsuperscript{4,5},
    Shuai Wang\textsuperscript{2,5,$*$}, 
    Xuelong Geng\textsuperscript{1}, Zhao Guo\textsuperscript{1}, Haoyu Li\textsuperscript{2,3}}, 
     \\ 
    Hao Yin\textsuperscript{4},  Xipeng Yang\textsuperscript{4}, Pengshen Zhang\textsuperscript{4}, Changwei Ma\textsuperscript{4}, Lei Xie\textsuperscript{1,}\thanks{Corresponding authors}
        \\[1.5ex]
        \\ Audio, Speech and Language Processing Group, Northwestern Polytechnical University\textsuperscript{1}
        \\ Nanjing University\textsuperscript{2}, Shanghai Jiao Tong University\textsuperscript{3}
        \\ GuaSemi Speech A Team\textsuperscript{4}, WeNet Community\textsuperscript{5},
}

\begin{document}

\maketitle

\begin{abstract}
    \label{sec:abs}

    In this paper, we present WEST(\textbf{WE} \textbf{S}peech \textbf{T}oolkit), a speech toolkit based on a large language model (LLM) for speech understanding, generation, and interaction.
    There are three key features of WEST:
    1) Fully LLM-based: Standing on the shoulders of giants by reusing mature architectures, ecosystems (e.g., Hugging Face), and methods (e.g., sequence packing) from large models.
    2) Full-stack: Supports tasks such as recognition, synthesis, understanding, dialogue, and multimodal capabilities, with extensibility to incorporate open-source models.
    3) Simple and Stupid: A simple and stupid speech toolkit that everyone can Touch.
    In addition, WEST provides two types of recipes, models, and experimental results.
    The first is entirely based on open-source models and open-source data,
    allowing users to fully reproduce the experiments in this paper and serving as a verification system or minimal system baseline. 
    The second is trained on massive data, offering superior performance so the user can directly apply it out of the box. WEST is publicly avilable at \url{https://github.com/wenet-e2e/west/}.
    \\[1.5ex]
    \textbf{Keywords:} \textit{Fully LLM based}, \textit{Full-stack}, \textit{Simple and Stupid}
    \end{abstract}

\section{Introduction}
\label{sec:intro}


Large language models (LLMs)\cite{radford2018improving, radford2019language, brown2020language, ouyang2022training} have had a profound impact on artificial intelligence(AI) and have become the core driving force in the field of AI today.
LLMs have changed the learning and application paradigms of AI. 
People can use LLMs through prompts, enhance the reasoning ability of models through CoT\cite{wei2022chain} methods like ReAct\cite{yao2023react}, use functions and tools, and further develop into Agent systems to handle complex tasks and scenarios.
In training, techniques such as mixture of experts models\cite{jiang2024mixtral}, Flash Attention\cite{dao2022flashattention}, 
and sequence packing\cite{krell2021efficient} have made training large models more efficient and economical,
while methods like deep think\cite{guo2025deepseek} have endowed LLMs with better understanding and reasoning capabilities.
In Inference, techniques such as quantization, speculative sampling\cite{chen2023accelerating},
and Paged Attention\cite{kwon2023vllm} have significantly improved the speed and efficiency of LLM inference.
In terms of ecosystem, an increasing number of models such as LLaMA\cite{touvron2023llama}, 
Mistral\cite{jiang2024mixtral}, DeepSeek\cite{bi2024deepseek}, and QWen\cite{bai2023qwen} are being open-sourced, making it easier for people to access the latest large models. Platforms like Hugging Face and ModelScope have also made it more convenient to use and share models.


Large models have also demonstrated powerful capabilities in Speech.
Models such as Whisper\cite{radford2023robust}, Seed-ASR\cite{bai2024seed},
SenseVoice\cite{an2024funaudiollm}, and Firered-ASR\cite{xu2025fireredasr}
have shown excellent performance in speech recognition tasks.
Models like VALL-E\cite{wang2023neural}, Seed-TTS\cite{anastassiou2024seed}, CosyVoice\cite{du2024cosyvoice}, 
and TouchTTS\cite{song2024touchtts} have made qualitative leaps in speech synthesis tasks compared to previous models,
with significant improvements in prosody, timbre, and emotion.
Speech understanding models such as QWen-Audio\cite{chu2023qwen}, and OSUM\cite{geng2025osum} can
not only recognize the text in speech but also understand the speaker's attributes such as emotion, gender, speech event and so on.
Models like GPT-4o\cite{hurst2024gpt}, Moshi\cite{defossez2024moshi}, Step-Audio\cite{huang2025step}, Freeze-Omni\cite{wang2024freeze},
GLM-4-Voice\cite{zeng2024glm}, Qwen2.5-Omni\cite{xu2025qwen2}, and OSUM-eChat\cite{geng2025osum} possess
end-to-end speech dialogue and multimodal capabilities, further enhancing the experience of speech interaction.
The aforementioned models have been open-sourced to varying degrees in terms of code, models, and data, greatly promoting the development of the speech field.


However, the aforementioned works have either open-sourced some of the inference code or models, but the training details and data of the models have not been open-sourced and are not accessible.
This prevents users from reproducing the training process of these models and conducting in-depth secondary development and research based on them.
Moreover, these models are mostly trained using different frameworks and internal data, making it difficult to conduct fair comparisons and evaluations.
Finally, these models mostly cover only one or a few tasks in speech, and there are significant differences between the publicly available models and codes for different tasks,
making task transfer and multi-task integration challenging.
We believe that the speech field still lacks a unified, full-stack, open-source, large model-based speech toolkit to better support research and applications in the speech domain.

In this paper, we present WEST, a speech toolkit based on LLM for speech understanding, generation, and interaction.
There are three key features of WEST:
\begin{itemize}
    \item \textbf{Fully LLM-based}: Standing on the shoulders of giants by reusing mature architectures, ecosystems (e.g., Hugging Face), and methods (e.g., sequence packing) from large models.
    \item \textbf{Full-stack}: Supports tasks such as recognition, synthesis, understanding, dialogue, and multimodal capabilities, with extensibility to incorporate open-source models.
    \item \textbf{Simple and Stupid}: A simple and stupid speech toolkit that everyone can Touch.
\end{itemize}

In addition, WEST provides two types of recipes, models, and experimental results. 
The first is entirely based on open-source models and open-source data,
allowing users to fully reproduce the experiments in this paper and serving as a verification system or minimal system baseline. 
The second is trained on massive data, offering superior performance so the user can directly apply it out of the box.

\section{WEST Toolkit}

\subsection{Data}
\label{sec:data}

\subsubsection{Data Format}
\label{sec:data_format}


Inspired by the training paradigm of text LLM, the training of large speech models typically consists of two stages: pre-training and fine-tuning.
The pre-training stage usually utilizes large-scale datasets, while the fine-tuning stage employs smaller-scale datasets.
Typically, massive foundational speech data may contain speech and optional text annotations.
If the foundational data has corresponding text labels, some tasks use speech recognition and synthesis for pre-training, such as QWen2.5-Omni, Freeze-Omni, OSUM-eChat, etc.
Other tasks use self-supervised learning with speech tokens for pre-training, such as Moshi, Step-Audio, etc.
In the fine-tuning stage, a small amount of high-quality annotated data with more information and various task types is usually used.



To support the aforementioned two training stages and consider data storage and reading efficiency, we have designed the following two data formats:

\begin{itemize}
    \item \textbf{Pre-training Data Format:} This data contains only speech and optional text labels, recorded in jsonl format. Each json includes two fields, `wav` and `txt`, corresponding to the speech file path and text label, respectively.
    Pre-training typically requires more than a million hours of data, containing over a billion speech files.
    To efficiently store and read massive foundational data, inspired by WeNet\cite{zhang2022wenet}, we also support packaging multiple foundational data entries using the tar format.
    During training, only the list paths of all tar packages need to be provided. WEST will download and extract the tar packages on-the-fly and read the speech and text data within them.
    Data compression and packaging not only greatly improve data storage and reading efficiency but also eliminate the need to record the addresses of all audio files, thereby saving memory during training. 
    See Listing \ref{lst:pretrain-jsonl} and \ref{lst:pretrain-tar} for examples of jsonl and tar formats, respectively.
    \item \textbf{Fine-tuning Data Format:} We use the role-content format commonly used in the large model field. 
    This format conveniently supports multi-turn interactions between users and systems and can be flexibly extended to modalities,
    which is crucial for tasks such as speech understanding, dialogue, and multi-turn interactions. 
    We save all fine-tuning data in jsonl format.
    See Listing \ref{lst:sft} for an example of role-content based data.
    The \textit{content} field can be either a text, an audio or mixed content.
\end{itemize}

\begin{cardlisting}[label={lst:pretrain-jsonl}]{Example: jsonl for Pre-training}
{"wav": "path/to/your/audio1.wav", "txt": "your text1 here"}
{"wav": "path/to/your/audio2.wav", "txt": "your text2 here"}
...
\end{cardlisting}

\begin{cardlisting}[label={lst:pretrain-tar}]{Example: tar list for Pre-training}
path/to/your/data1.tar
path/to/your/data2.tar
...
# Note: each tar file contains multiple wavs and txts.
\end{cardlisting}

\begin{cardlisting}[label={lst:sft}]{Example of role-content based data for Fine-tuning}
{
   "messages":[
      {
         "role":"user",
         "content": "your text1 here"
      },
      {
         "role":"assistant",
         "content":{
            "type":"audio",
            "audio":"path/to/your/audio2.wav",
            "text":"your text2 here"
         }
      },
    {
         "role":"user",
         "content":{
            "type":"audio",
            "audio":"path/to/your/audio3.wav",
            "text":"your text3 here"
         }
      },
      {
         "role":"assistant",
         "content": "your text4 here"
      }
   ]
}

# Note: Here we just unwrap one json line for better readability.
\end{cardlisting}

\subsubsection{Data packing}

Training large language models (LLMs) is a computationally demanding task. 
It requires vast amounts of data, powerful hardware, and clever optimization techniques. 
Naive batching can lead to significant padding tokens, especially when dealing with variable-length sequences, which is common in speech data.
Previously, data was typically sorted by length and then assigned to different batches, which can reduce the amount of padding.
Alternatively, dynamic batch sizes were used, adjusting the batch size based on the length of the longest sequence in each batch.
However, dynamic batch sizes can lead to inconsistent computation and memory usage across batches, affecting training stability and efficiency.

Sequence packing is a more efficient way to handle variable-length sequence data and is widely used in the training of text LLMs.
Packed sequences offer an elegant solution. 
Instead of padding, we concatenate multiple shorter sequences into a single, longer sequence.
This minimizes wasted compute (through padding tokens). 
It also allows us to process more tokens per batch thus reducing training time.
As long as we ensure the model doesn’t attend across sequence boundaries, we can safely pack sequences.
Fortunately, the self-attention mechanism of Transformers naturally supports this.
Moreover, Flash attention\cite{dao2022flashattention} already has a very efficient implementation for this.

In WEST, we support batch and sequence packing for training for both pre-training and fine-tuning stages.
\ref{exp:data_pack} provides an experiment of the comparisons of static
 batch, dynamic batch and sequence pack.

\subsection{Models}
\label{sec:models}


WEST includes various built-in task models, such as speech recognition and understanding, speech synthesis, and speech dialogue.
Users can also quickly build their own models based on the basic components provided by WEST.
At the same time, WEST also includes popular open-source speech models that users can directly use for training and inference.

In this section, we will introduce the model design in WEST, including built-in models and open-source models.

\subsubsection{TouchASU: Speech Recognition and Understanding}

\begin{figure}[!ht]
    \centering
    \includegraphics[width=0.8\textwidth]{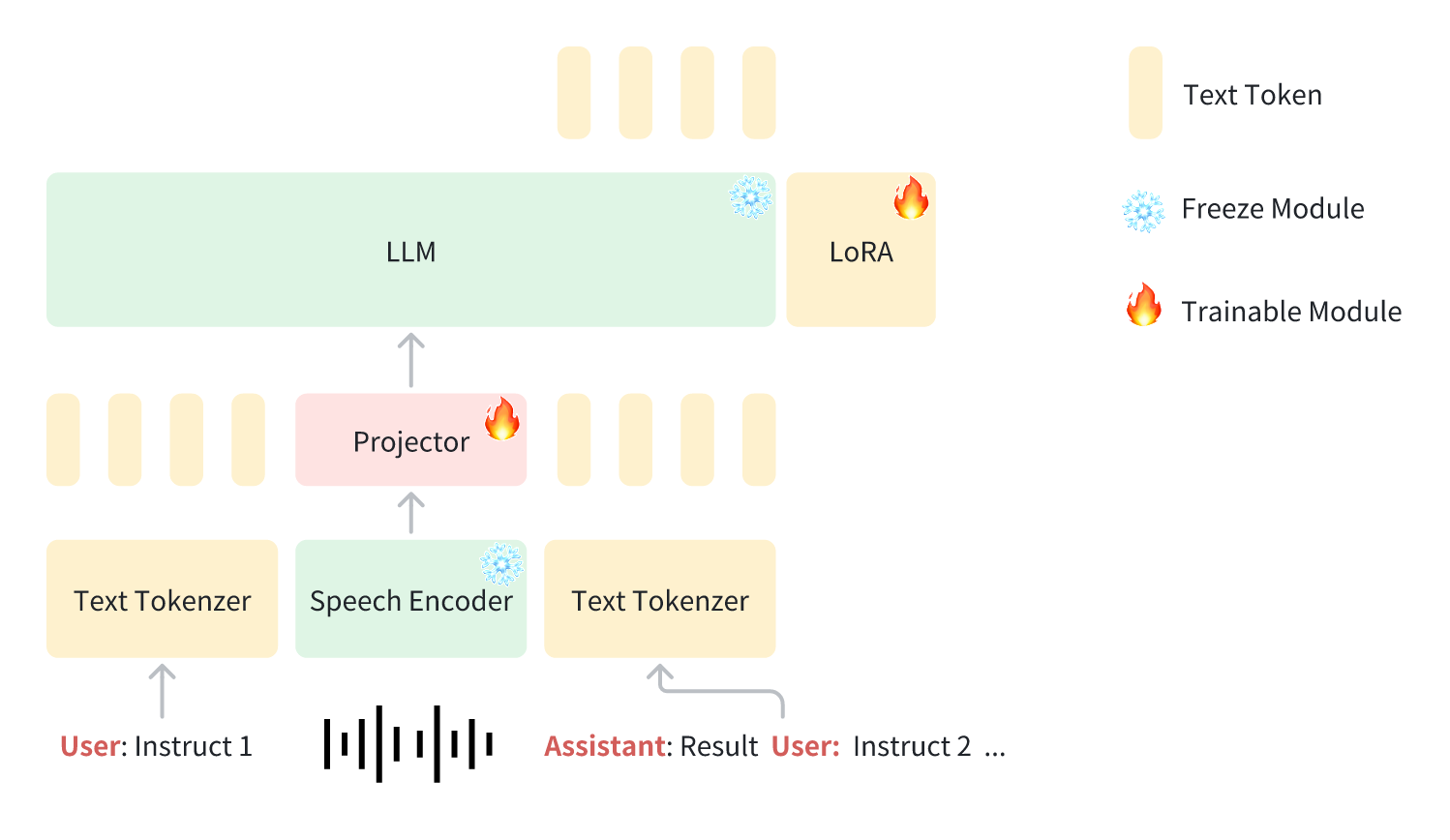}
    \caption{TouchASU for Speech Recognition and Understanding}
    \label{fig:touchasu}
\end{figure}

Inheriting our previous work TouchASP\cite{song2024touchasp}, the TouchASU model is designed in WEST to support speech recognition, speech understanding,
and Speech Question Text Answering (SQTA) tasks.
As shown in the Fig \ref{fig:touchasu}, TouchASU consists of a Speech Encoder, a Projector, and an LLM,
with an optional configurable Low-Rank Adaptation(LoRA) for the LLM part.

We use WeNet\cite{yao2021wenet,zhang2022wenet} as the Speech Encoder, which supports several SoTA open-source models, 
such as Whisper\cite{radford2023robust}, FireRed\cite{xu2025fireredasr}, Paraformer\cite{gao2022paraformer}, etc.
Users can flexibly choose suitable pre-trained models according to their tasks and needs.
For example, for Chinese speech recognition tasks, models like FireRed and Paraformer that perform better in Chinese can be selected.
For multilingual speech recognition and translation tasks, the Whisper model with better multilingual support can be chosen.
Compared to the original Whisper, in WeNet, we do not need to pad the data to a fixed length of 30 seconds,
but instead use the actual length of the data for training and inference, greatly improving the efficiency of training and inference.

As SLAM-ASR\cite{ma2024embarrassingly}, we use a combination of Convolution + Linear as the Projector to map the features output by the Speech Encoder to the input dimension of the LLM.
Users can control the downsampling rate of the Projector by adjusting the stride of the Convolution,
thus balancing performance and computational overhead.

Typically, we freeze the parameters of the LLM and Speech Encoder, only training the Projector and the LoRA part of the LLM.
However, users can flexibly choose the training strategy based on the characteristics of their tasks,
such as fine-tuning all parameters of the LLM or fine-tuning the parameters of the Speech Encoder.

As illustrate in \ref{sec:data_format}, TouchASU has built-in support for multi-turn and multimodal interactions,
making it highly suitable for mixed-modal, multi-turn Speech QA.

\subsubsection{TouchTTS: Text to Speech}

\begin{figure}[!ht]
    \centering
    \includegraphics[width=1.0\textwidth]{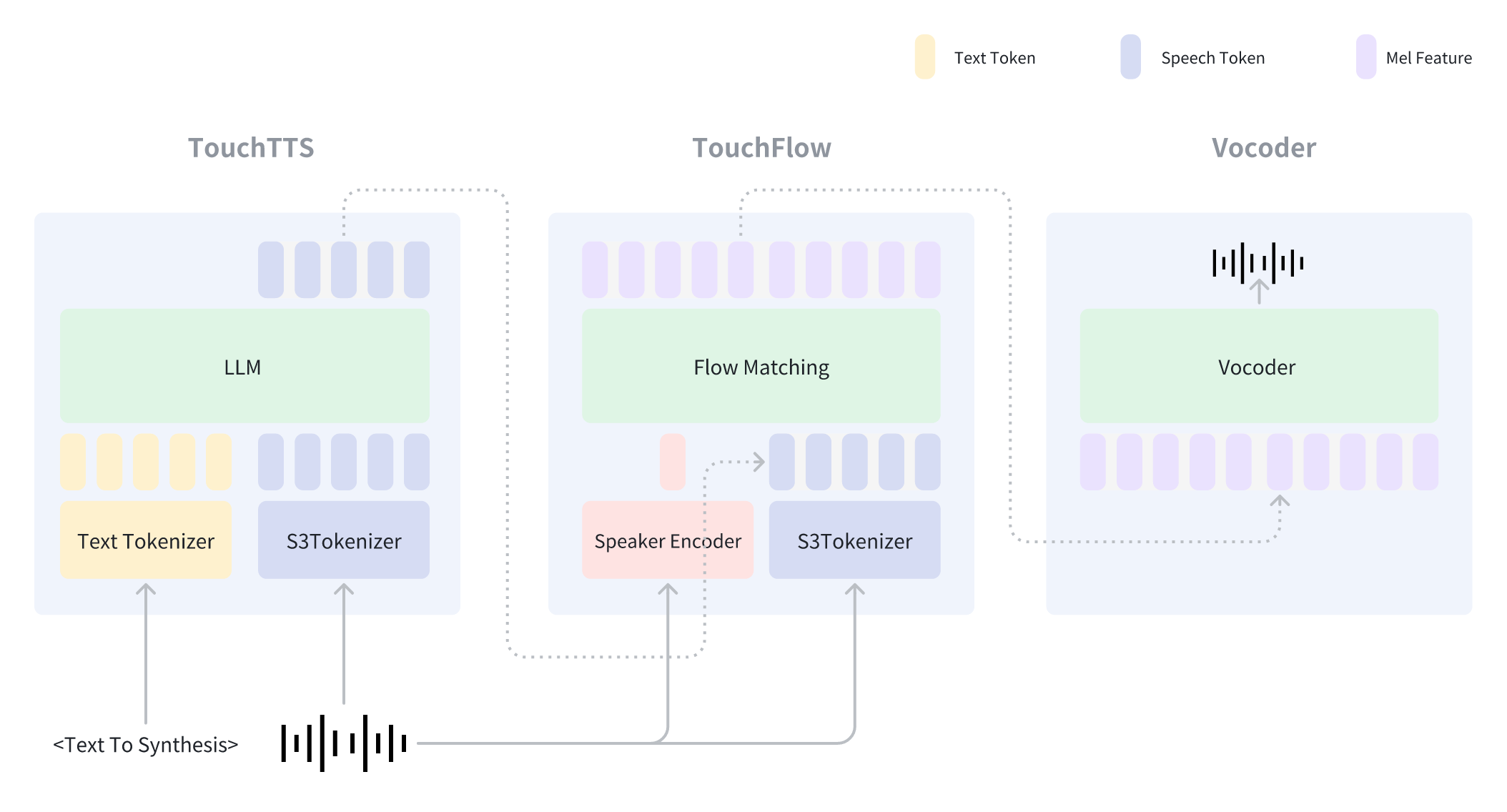}
    \caption{TouchTTS for Text to Speech, the dashed line indicates the inference data flow.}
    \label{fig:touchtts}
\end{figure}

As TTS task, our model design is almost identical to our previous work, TouchTTS \cite{song2024touchtts}.
As shown in Fig \ref{fig:touchtts}, our TTS system consists of three components: TouchTTS, TouchFlow, and Vocoder.
TouchTTS is used for modeling from text to speech tokens.
TouchFlow is a flow matching model that converts speech tokens into mel-spectrogram features.
The Vocoder then converts the mel-spectrogram into PCM samples.
We use S3Tokenizer \cite{du2024cosyvoice} for speech token extraction,
and wespeaker \cite{wang2023wespeaker} is applied as the speaker encoder.

Our design principle is to leverage the LLM ecosystem as much as possible,
so both TouchTTS and TouchFlow are based on the LLM backbone.
We also use the same text tokenizer as LLM, so we can reuse the pre-trained LLM weights.

\subsubsection{TouchChat: Thinker Talker Speech Chat}

\begin{figure}[!ht]
    \centering
    \includegraphics[width=1.0\textwidth]{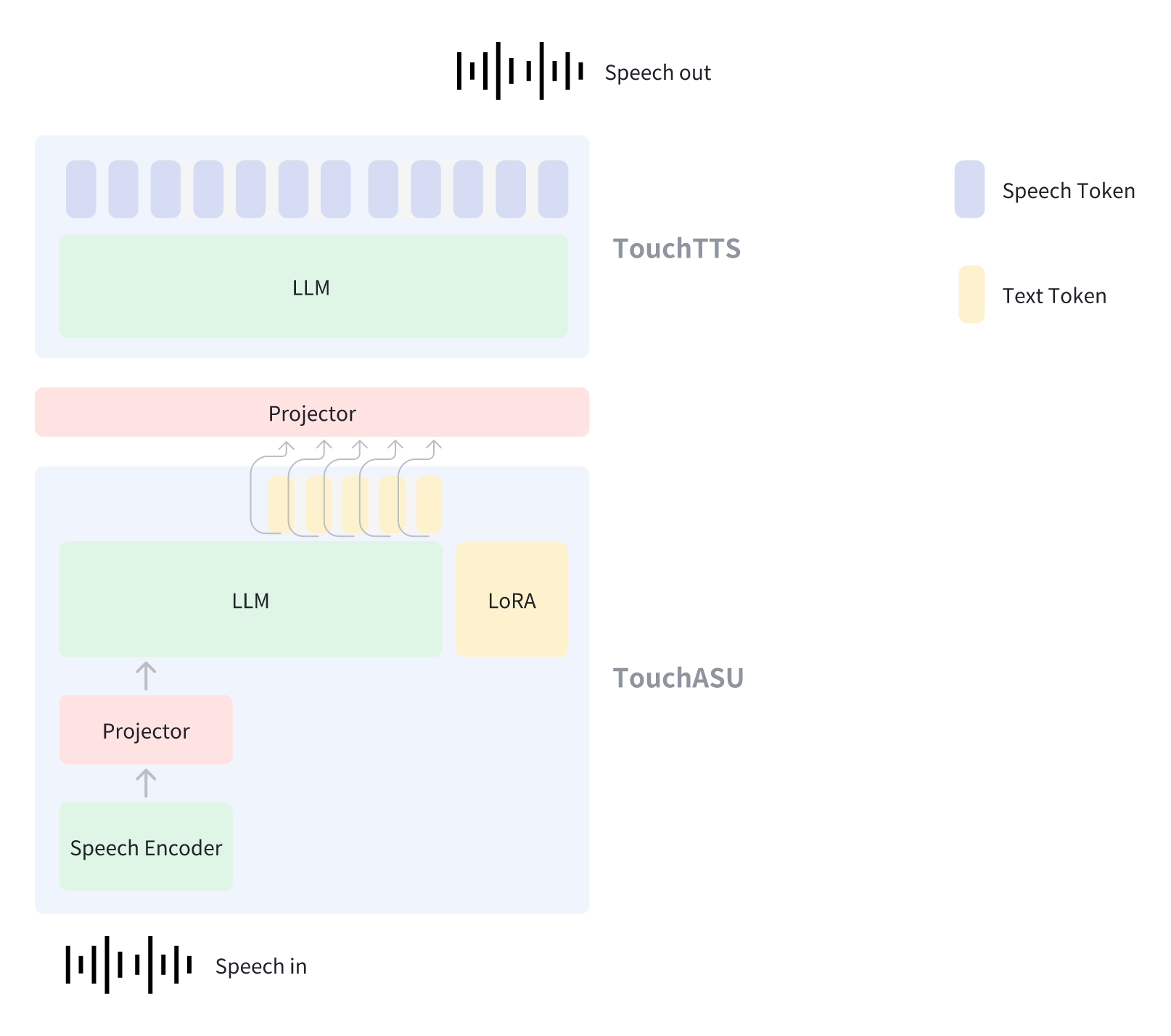}
    \caption{TouchChat pipelined by TouchASU and TouchTTS}
    \label{fig:touchchat}
\end{figure}


To build end-to-end speech interaction systems, we refer to Qwen2.5-Omni\cite{xu2025qwen2}, we design the TouchChat model.
TouchChat consists of TouchASU and TouchTTS, and TouchASU is used to understand user speech, and TouchTTS is used to speech synthesis.
Two modules are connected by a Projector, the input of the Projector is the output of the last embedding layer of TouchASU,
and the output of the Projector is used as the input of TouchTTS.
By this way, TouchChat can effectively use the pre-trained understanding and synthesis models, and quickly build end-to-end dialogue systems at a low cost.
At the same time, the input of TouchTTS also contains the embedding representation of user speech and the embedding representation of the understanding results of TouchASU,
so that more rich context information can be used to further improve the performance of TouchTTS.
In addition, we can also freeze the parameters of TouchASU to avoid lowering the understanding ability during synthesis training.

\subsubsection{TouchChat2: All in One Speech Chat}


TouchChat2 is an all in one speech chat system.
TouchChat2 integrates understanding ability and synthesis ability into a single model and trains end-to-end dialogue ability based on it.
Compared to TouchChat, TouchChat2 has a simpler structure, is more end-to-end, and also has higher system upper bound,
at the cost of more data, more computing resources, and more detailed model tuning.

\subsubsection{TouchOmni}

\subsubsection{Open-source Models}

Currently, we support OSUM-echat\cite{geng2025osum}, you can finetune the models in WEST
out-of-the-box.

\section{Experiments}
\label{sec:experiments}

In this section, we show the results of WEST on speech recognition, speech understanding, speech synthesis, and end-to-end speech chat.
First, we will give the standard open source models and datasets, and then give the results on large-scale datasets.
Users can precisely reproduce the standard datasets and models, 
and all the recipes on standard datasets are publicly available in our official repository.
We also show the results of sequence pack.

\subsection{Speech Recognition}


We give the standard reference results on speech recognition on AIShell\cite{bu2017aishell} and LibriSpeech\cite{panayotov2015librispeech}.
We vary the combination of LLM, Speech Encoder, and LoRA on AIShell.
As shown in Table \ref{table:asr}, WEST shows promising results on speech recognition tasks.

\begin{table}[!ht]
\centering
\label{table:asr}
\caption{Speech Recognition Results on WEST}
\begin{tabular}{@{}llllll@{}}
\toprule
Dataset                      & LLM                       & Speech Encoder                          & LoRA               & test set    & WER/CER \\ \midrule
\multirow{3}{*}{AIShell}     & Qwen3-1.7B                & firered                                 & Y                  & test        & 4.17    \\
                             & Qwen2-7B                  & firered                                 & N                  & test        & 4.01    \\
                             & Qwen2-7B                  & paraformer                              & Y                  & test        & 3.51    \\ \midrule 
\multirow{2}{*}{LibriSpeech} & \multirow{2}{*}{QWen2-7B} & \multirow{2}{*}{whisper-large-v3-turbo} & \multirow{2}{*}{Y} & test\_clean &         \\
                             &                           &                                         &                    & test\_other &         \\ \bottomrule 
\end{tabular}
\end{table}

\subsection{Speech Understanding and Question Answering}


As speech understanding, we train a question answering task on Belle-1.4M-SLAM-Omni\cite{chen2024slam}.
The training is divided into two stages.
In the first stage, we train on AIShell-2\cite{du2018aishell} for speech recognition.
In the second stage, we train on Belle-1.4M-SLAM-Omni for question answering.
We also add AIShell-2 data in the second stage to avoid the loss of speech recognition ability.
We apply the firered model as the speech encoder and Qwen3-1.7B as the LLM,
and the model is trained on 4 A800 GPUS with pack size 25000 and 20000 steps. During the testing stage, we use beam search decoding. For detailed configuration, please refer to \url{https://github.com/wenet-e2e/west/blob/main/examples/belle_1.4M_qa/conf/generation_config.json}.


We also design a Chinese question answering test set with 300 samples,
which covers traditional Chinese culture, geography, STEM(Science, Technology, Engineering, Mathematics) knowledge.
We name it Chinese QA test set, which is publicly available at \url{https://huggingface.co/datasets/wenet-e2e/chinese_qa}.
We evaluate the question answering ability of models on this test set.

Table\ref{table:qa} show the results on speech question answering on WEST.

\begin{table}[!ht]
\centering
\label{table:qa}
\caption{Speech Question Answering Results on WEST}
\begin{tabular}{@{}llllll@{}}
\toprule
Dataset                                & LLM                         & LoRA               & test                  & Result  \\ \midrule
\multirow{2}{*}{Belle\_1.4M-SLAM-Omni} & \multirow{2}{*}{Qwen3-1.7b} & \multirow{2}{*}{Y} & AIShell-2 test (WER)  & 4.49 \% \\
                                       &                             &                    & Chinese QA test (ACC) & 70.0\%  \\ \midrule
\end{tabular}
\end{table}

\subsection{Speech Synthesis}

LibriTTS\cite{zen2019libritts} is used as our speech synthesis dataset.
We use QWen-0.5B as the base model for TouchTTS,
and we add 4096 speech tokens in the tokenizer.
The TouchTTS model is trained on 8 A800 GPUs with pack 20000, 40000 steps.
The TouchFlow is trained on 8 3090 GPUs with batch 64, 50000 steps.
We use Whisper for WER evaluation and wespeaker for Speaker Simlarity(SS) evaluation.
Table\ref{table:tts} shows the results on speech synthesis on WEST.

\begin{table}[!ht]
\centering
\label{table:tts}
\caption{Speech Synthesis Results on WEST}
\begin{tabular}{@{}lllll@{}}
\toprule
Dataset  & TouchLLM   & TouchFlow  & WER  & SS    \\ \midrule
LibriTTS & Qwen2-0.5B & Qwen2-0.5B & 5.15 & 0.847 \\ \bottomrule
\end{tabular}
\end{table}

\subsection{Speech Chat}
We trained TouchChat model using the AIShell-2, Wenetspeech4TTS and Belle-1.4M-SLAM-Omni datasets. The training is divided into three stages. In the first stage, we train TouchASU model using AIshell-2 and Belle-1.4M-SLAM-Omni datasets. In the second stage, we train TouchTTS model using Wenetspeech4TTS dataset. In the third stage, we train TouchChat using Belle-1.4M-SLAM-omni, which TouchASU is freezed. The token2wav module is based on CosyVoice-300M-SFT \footnote{https://www.modelscope.cn/models/iic/CosyVoice-300M-SFT/}, which converts the discrete speech tokens generated by the TouchChat into continuous audio waveforms. During the testing phase, the thinker module uses greedy decoding, while the talker module uses sampling decoding.
We evaluate the spoken question answering capability of TouchChat on Chinese QA test set. Two evaluation methods are employed: S2T, where the text responses generated by the model are evaluated directly, ans S2S, where the model's speech responses are transcribed using Paraformer \footnote{https://www.modelscope.cn/models/iic/speech\_paraformer-large\_asr\_nat-zh-cn-16k-common-vocab8358-tensorflow1} before evaluation. 

Table \ref{table:chat} show the results on speech chat on WEST.

\begin{table}[!ht]
\centering
\label{table:chat}
\caption{Speech Chat Results on WEST}
\begin{tabular}{lllll}
\toprule
\multirow{2}{*}{Dataset} & \multirow{2}{*}{Thinker} & \multirow{2}{*}{Talker} & \multicolumn{2}{l}{Chinese QA test} \\
                         &                          &                         & S2T(ACC)              & S2S(ACC)              \\ \midrule
Belle\_1.4M-SLAM-Omni    & Qwen3-1.7B               & Qwen2-0.5B              & 64.66\%          & 54.33\%          \\ \bottomrule
\end{tabular}
\end{table}

\subsection{Data pack}
\label{exp:data_pack}

The above experiments are all based on sequence pack training,
which demonstrates the correctness of Sequence pack in WEST.
In this section, we conduct a simple experiment to illustrate the advantages of sequence pack.
We train an ASR model on an RTX 3090 using WEST, with a 40M Conformer as the Speech Encoder,
QWen-1.5B as the LLM, and AIshell-1 as the training data.
We compare three training methods: static batch, dynamic batch, and sequence pack.
In each method, we maximize GPU memory usage without causing OOM,
and compare the time taken to process 10,000 utterances.


As shown in Table \ref{table:pack}, sequence pack achieves higher GPU utilization and faster training time without causing OOM.
In the experimental task, compared to static batch,
sequence pack improves training speed by 2.4 times. 

\begin{table}[!ht]
\centering
\label{table:pack}
\caption{Comparison of different training methods on training time and GPU utilization.}
\begin{tabular}{@{}llll@{}}
\toprule
methods       & Best Configuration       & GPU SM utils(\%) & time on training 10000 utts \\ \midrule
static batch  & batch size 32            & 63.05            & 9 min 33s                   \\
dynamic batch & max token in batch 4096  & 71.49            & 6 min 28s                   \\
sequence pack & Pack token 8192          & 73.87            & 3 min 58s                   \\ \bottomrule
\end{tabular}
\end{table}

\section{Limitations and Future Works}
\label{sec:futurework}


Currently, WEST is still under rapid iterative development. We will continue to update this paper until WEST reaches a stable 1.0 version.
Please stay tuned for updates on WEST.

\bibliographystyle{unsrt}
\bibliography{refs}

\end{document}